\documentclass{article}
\PassOptionsToPackage{numbers, compress}{natbib}

\usepackage[preprint]{neurips_2026}

\usepackage[utf8]{inputenc} % allow utf-8 input
\usepackage[T1]{fontenc}    % use 8-bit T1 fonts
\usepackage{hyperref}       % hyperlinks
\usepackage{url}            % simple URL typesetting
\usepackage{graphicx}       % figures
\usepackage{capt-of}       % captions outside floats
\usepackage{booktabs}       % professional-quality tables
\usepackage{amsfonts}       % blackboard math symbols
\usepackage{amssymb}        % additional math symbols
\usepackage{amsmath}        % math environments
\usepackage{nicefrac}       % compact symbols for 1/2, etc.
\usepackage{microtype}      % microtypography
\usepackage[table]{xcolor}  % table colors
\usepackage{xspace}
\usepackage{pifont}
\usepackage{multirow}
\usepackage{threeparttable}
\usepackage{enumitem}
\usepackage{pdflscape}

\setlist[itemize]{leftmargin=1.6em,itemsep=0.1em,topsep=0.2em,parsep=0pt,partopsep=0pt}
\setlist[enumerate]{leftmargin=1.7em,itemsep=0.1em,topsep=0.2em,parsep=0pt,partopsep=0pt}
\makeatletter
\renewcommand{\paragraph}{%
  \@startsection{paragraph}{4}{\z@}%
                {0.7ex \@plus 0.25ex \@minus 0.18ex}%
                {-1em}%
                {\normalsize\bf}%
}
\makeatother
\newcommand{\ourdataset}{{\textsc{\textbf{SeePhys Pro}}}\xspace}

\definecolor{FigOneDeepRed}{RGB}{210,92,70}
\definecolor{FigOneSoftRed}{RGB}{232,130,112}
\definecolor{FigOneLightYellow}{RGB}{210,178,74}
\definecolor{FigOneDeepGreen}{RGB}{54,150,112}
\definecolor{FigOneSoftGreen}{RGB}{104,188,140}
\definecolor{FigOneDeepBlue}{RGB}{48,112,184}
\definecolor{FigOneSoftBlue}{RGB}{94,152,210}
\definecolor{FigOnePaleBlue}{RGB}{132,180,224}

\newcommand{\ourdatasettitle}{%
  {\textsc{%
    \textbf{%
      \textcolor{FigOneDeepRed}{S}%
      \textcolor{FigOneSoftRed}{e}%
      \textcolor{FigOneLightYellow}{e}%
      \textcolor{FigOneDeepGreen}{P}%
      \textcolor{FigOneSoftGreen}{h}%
      \textcolor{FigOneSoftBlue}{y}%
      \textcolor{FigOnePaleBlue}{s}%
      \ % space between SeePhys and Pro
      \textcolor{FigOneDeepBlue}{P}%
      \textcolor{FigOneSoftBlue}{r}%
      \textcolor{FigOnePaleBlue}{o}%
    }%
  }}%
  \xspace%
}

\definecolor{improvegreen}{RGB}{0, 128, 0} 
\definecolor{improvered}{RGB}{255,0,0}

\definecolor{delectricred}{RGB}{255,0,0}
\colorlet{lightdelectricred}{delectricred!30}

\definecolor{delectricgreen}{RGB}{0,255,0}
\colorlet{lightdelectricgreen}{delectricgreen!30}

\hypersetup{
    colorlinks=true,      
    urlcolor=magenta,
}

\title{
\ourdatasettitle: Diagnosing Modality Transfer and Blind-Training Effects in Multimodal RLVR for Physics Reasoning
}

\author{
\textbf{Kun Xiang$^1$, Terry Jingchen Zhang$^2$, Zirong Liu$^1$, Bokai Zhou$^1$,} \\
\textbf{Yueling Tang$^1$, Junjie Yu$^1$, Jiacong Lu$^1$, Shangrui Huang$^1$, Heng Li$^1$, Likui Zhang$^1$,} \\
\textbf{Kunkun Liu$^4$, Changzheng Zhang$^4$, Yangle Fang$^4$, Boqiang Guo$^4$, Hui-Ling Zhen$^4$,} \\
\textbf{Dandan Tu$^{4,*}$, Yinya Huang$^{2,3}$, Xiaodan Liang$^{1,*}$} \\
\\
$^{1}$Sun Yat-sen University \quad
$^{2}$ETH Zurich \quad
$^{3}$ETH AI Center \quad
$^{4}$Huawei Technologies Ltd. \\
$^{*}$Corresponding authors.
}

\begin{document}

\maketitle

\begin{abstract}
We introduce \ourdataset, a fine-grained modality transfer benchmark that studies whether models preserve the same reasoning capability when critical information is progressively transferred from text to image.
Unlike standard vision-essential benchmarks that evaluate a single input form, \ourdataset features four semantically aligned variants for each problem with progressively increasing visual elements.
Our evaluation shows that current frontier models are far from representation-invariant reasoners: performance degrades on average as information moves from language to diagrams, with visual variable grounding as the most critical bottleneck.
Motivated by this inference-time fragility, we further develop large training corpora for multimodal RLVR and use blind training as a diagnostic control, finding that RL with all training images masked can still improve performance on unmasked validation sets.
To analyze this effect, text-deletion, image-mask-rate, and format-saturation controls suggest that such gains can arise from residual textual and distributional cues rather than valid visual evidence.
Our results highlight the need to evaluate multimodal reasoning not only by final-answer accuracy, but also by robustness under modality transfer and by diagnostics that test whether improvements rely on task-critical visual evidence. 

\textbf{Project Page: }\url{https://seephyspro.github.io}.

\textbf{Challenge: }\url{https://www.codabench.org/competitions/16010/}.

\textbf{GitHub: }\url{https://github.com/AI4Phys/SeePhy-Pro}.

% \textbf{Project Page and Data:} An anonymized project page and data repository will be provided for review.

\end{abstract}

\section{Introduction}

A key challenge for multimodal AI is \emph{modality consistency}, namely whether a model preserves the same reasoning behavior when equivalent information is expressed in different forms~\cite{MathVerse,MathVista,MMMU,CrossMath}. This gap is easy to miss when benchmarks evaluate a single input format, and improvements in final-answer accuracy do not necessarily imply representation-invariant reasoning. Physics provides a particularly sharp testbed, since a diagram can define the physical system itself rather than merely illustrate the text~\cite{ScienceQA,SeePhys,PhyX,PhysReason,PhysicsArena}. Here, \emph{structure} refers to the schema of the system, such as circuit connectivity in a circuit diagram, the contact graph and force directions in a mechanics sketch, or the topology of optical elements in a ray diagram. \emph{Variables} refer to the labeled quantities tied to specific entities or relations, such as voltages and currents attached to particular nodes and branches, masses tied to specific blocks, or angles tied to specific rays. As information shifts from text into vision, the model must perform grounding and semantic binding, not just generic perception.

To address this gap, we introduce \ourdataset, a fine-grained modality-transfer benchmark built on the principle of \emph{same physics, different representation}. Each problem has four aligned variants that progressively move task-critical information from language to vision: (L1) text-only, (L2) structure-in-image, (L3) structure+variables-in-image, and (L4) fully rendered problem image. This setup decomposes performance degradation into structural transfer, variable grounding, and full-rendering effects. Across a wide range of MLLMs, average performance drops as information is transferred from text to vision, with the largest degradation often occurring when variables must be grounded from the image.
We have also released \ourdataset as Challenge~3\footnote{\url{https://www.codabench.org/competitions/16010/}.} in the 3rd AI for Math Workshop at ICML 2026.\footnote{\url{https://ai4math2026.github.io/}.}

To study training-time behavior, we further build two large-scale multimodal RLVR corpora, PhysRL-38K and PhysRL-8K. While recent multimodal RLVR studies improve visual reasoning performance~\cite{PAPO,VisualRFT,MMEureka,R1Onevision}, outcome-only rewards may still encourage shortcuts that do not depend on valid visual evidence~\cite{GroundedCoT,MoreThinkingLessSeeing,LimitRLVR}. We therefore include a blind-training control that masks all training images, making each training instance visually unsolvable. Surprisingly, this blind-training RL still improves accuracy on unmasked validation sets, showing that models can infer or reconstruct useful reasoning paths from unsolvable text-only inputs. Further text-deletion and mask-rate controls suggest that these gains are likely driven by residual language, problem templates, and dataset-level statistical regularities rather than effective visual evidence. Taken together, the training-time and test-time results highlight the need for future multimodal reasoning research to look beyond absolute accuracy gains and examine whether such gains come from task-critical visual evidence or from shortcuts in textual structure.

Our contributions are summarized as follows:
\begin{itemize}
    \item We introduce \ourdataset, a progressive modality-transfer benchmark grounded in multimodal physics reasoning, together with metrics that decompose performance into structure recognition, variable grounding, modality gap, and representation consistency.
    \item We evaluate a wide range of closed- and open-weight MLLMs and find that even frontier models remain fragile under modality transfer, with the largest degradation often occurring when variables must be extracted from image(s).
    \item We release PhysRL-38K and PhysRL-8K as source-matched, test-disjoint physics RL training corpora, and use blind training (masked-image RL) as a negative-control setting; we find it can still improve unmasked test accuracy without reliably closing modality-transfer gaps, showing that accuracy gains do not necessarily imply visually grounded learning.
\end{itemize}

\section{Related Work}
\label{sec:related}

\paragraph{Multimodal physics reasoning.}
General science and expert reasoning benchmarks such as ScienceQA~\cite{ScienceQA}, MMMU~\cite{MMMU}, and OlympiadBench~\cite{OlympiadBench} show that multimodal scientific problem solving remains difficult. More specialized physics benchmarks, including SeePhys~\cite{SeePhys}, PhyX~\cite{PhyX}, PhysReason~\cite{PhysReason}, PhysicsArena~\cite{PhysicsArena}, and QuantiPhy~\cite{QuantiPhy}, further highlight the difficulty of diagram-based physical reasoning. However, most evaluate each problem in a fixed input form. \ourdataset instead studies a controlled modality-transfer setting where the underlying physics is fixed while structure, variables, and the full statement are progressively moved from text into vision.

\paragraph{Vision grounding ability in reasoning.}
Several benchmarks test whether MLLMs truly use visual evidence during reasoning. MathVista~\cite{MathVista}, MathVerse~\cite{MathVerse}, and CrossMath~\cite{CrossMath} use visual mathematical problems or information-controlled variants to expose modality gaps. Our setting is stricter: physics diagrams often define the system itself (e.g., topology and variable-to-entity bindings), and \ourdataset separates structural grounding, variable grounding, and full-rendering effects across aligned levels.

\paragraph{Reinforcement learning with verifiable rewards.}
Reinforcement learning with verifiable rewards (RLVR) is widely used to post-train reasoning models~\cite{GRPO,DAPO,GSPO}, and has also been explored in multimodal settings~\cite{PAPO,VLRethinker,VisualRFT,MMEureka,R1Onevision}. However, outcome-only rewards may encourage shortcuts that do not depend on valid visual evidence, and recent analyses suggest that reasoning-style training can amplify ungrounded behavior or improve behaviors already latent in the base model~\cite{GroundedCoT,MoreThinkingLessSeeing,LimitRLVR}. We therefore use blind training (masking all training images) as a negative control: if RL still improves unmasked test performance, the gain is not fully attributable to better visual grounding.

\section{\ourdataset: A Fine-Grained Benchmark for Modality-Transfer}
\label{sec:benchmark}

\paragraph{Design principle.}
\ourdataset is built around the diagnostic principle of \emph{same physics, different representation}.
For each seed problem, the physical system, required laws, answer, and reasoning target are fixed, while problem-critical information is progressively moved across modalities.
This design turns a single physics question into a controlled probe of whether MLLMs reason over stable physical semantics or over surface-level input formats, following controlled-modality diagnostics from visual mathematics and modality-gap evaluation~\cite{MathVerse,CrossMath}.
It also addresses an ambiguity in ordinary vision-essential physics evaluation~\cite{SeePhys,PhyX,PhysReason}: when a model succeeds or fails on a visual physics question, final accuracy alone cannot tell whether the decisive factor is diagram perception, variable reading, physical abstraction, OCR, or downstream symbolic reasoning.

\begin{figure*}[t]
    \centering
    \includegraphics[width=0.98\textwidth]{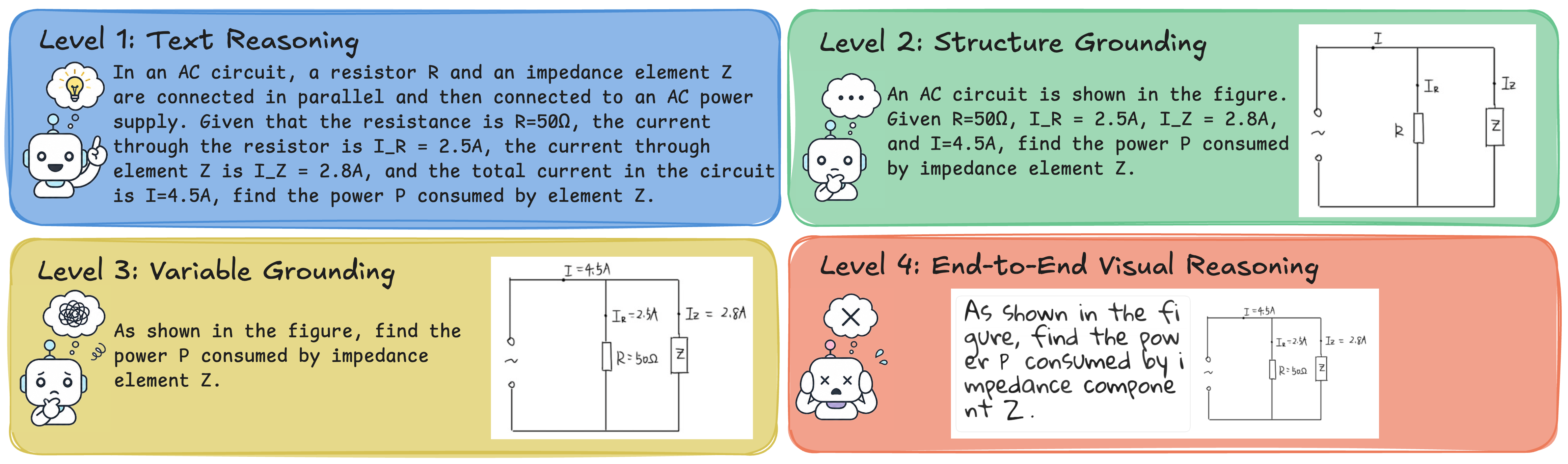}
    \vspace{-0.8em}
    \caption{\textbf{Overview of the four modality-transfer levels in \ourdataset.}
    Each seed problem is transformed into four semantically aligned variants that progressively move problem-critical information from language to vision: Level~1 is text-only, Level~2 moves structure into the image, Level~3 further moves variables and labels into the image, and Level~4 renders the full problem into a single visual input.}
    \label{fig:bmk_overview}
    \vspace{-0.5em}
\end{figure*}

\paragraph{Four-level modality transfer.}
Figure~\ref{fig:bmk_overview} illustrates the four-level transformation with an example.
Each seed problem is converted into four aligned variants with identical physical semantics.
The variants are not independently written questions: annotators manually redraw and edit the same problem so that the physical system, queried quantity, variables, constraints, solution path, and gold answer remain unchanged.
Only the carrier of information changes across levels.
Level 1 is \emph{text-only}: all structural relations, variables, and numerical quantities are described in language, providing a reference point for text-based physics reasoning.
Level 2 adds structured visual information by moving only the physical structure into the image while keeping variables in text, testing visual structural understanding such as circuit topology, force configuration, graph layout, or pulley connection.
Level 3 further overlays variables and labels onto the same diagram, testing whether models can read quantities and bind them to the correct physical entities.
Level 4 builds upon Level 3 by converting the problem statement into handwritten text and rendering it alongside the diagram into a single image. This forces the model to simultaneously process handwritten formulas, complex layouts, and physical reasoning within a unified visual context.
This controlled construction makes Level 1--4 a modality-transfer probe rather than a collection of different problems.

\paragraph{Benchmark data collection.}
We collect seed problems from heterogeneous physics sources rather than directly reusing existing fixed-form physics benchmarks~\cite{SeePhys,PhysReason,OlympiadBench,PhyX}, including public datasets, textbooks and problem books, PhD qualifying and entrance examinations, olympiad archives, and school or university exam papers. The source pool contains over 5,000 PDF pages, which are processed with Mathpix OCR~\cite{mathpix} and then curated by 10 engineering-trained annotators, including 7 bachelor's-level and 3 PhD-level annotators. Each accepted problem is assigned a three-level taxonomy covering discipline, field, and domain.
During curation, we filter invalid samples, normalize notation and answer formats, and remove near-duplicates using script-based text matching, manual review, and GPT-5-mini~\cite{openai_gpt5_systemcard} for LLM-assisted checks. 
During transformation, annotators rewrite accepted seeds into four aligned variants while preserving the physical system, target quantity, and gold answer. Diagrams are then redrawn and separated into structure and variable layers, enabling Level~2 to test structure grounding and Level~3 to test variable grounding. We filter out problems with uncertain answers, incomplete statements, or insufficient solution conditions, and verify that each four-level group preserves the same physical system, variables, answer, and reasoning path.
The full construction workflow is described in Appendix~\ref{app:construction_details}.

\begin{figure*}[t]
    \centering
    \vspace{0.45em}
    \begin{minipage}[t]{0.3\textwidth}
        \centering
        \caption{\textbf{3-levels taxonomy.}}
        \label{fig:taxonomy_distribution}
        \vspace{0.2em}
        \includegraphics[width=\textwidth]{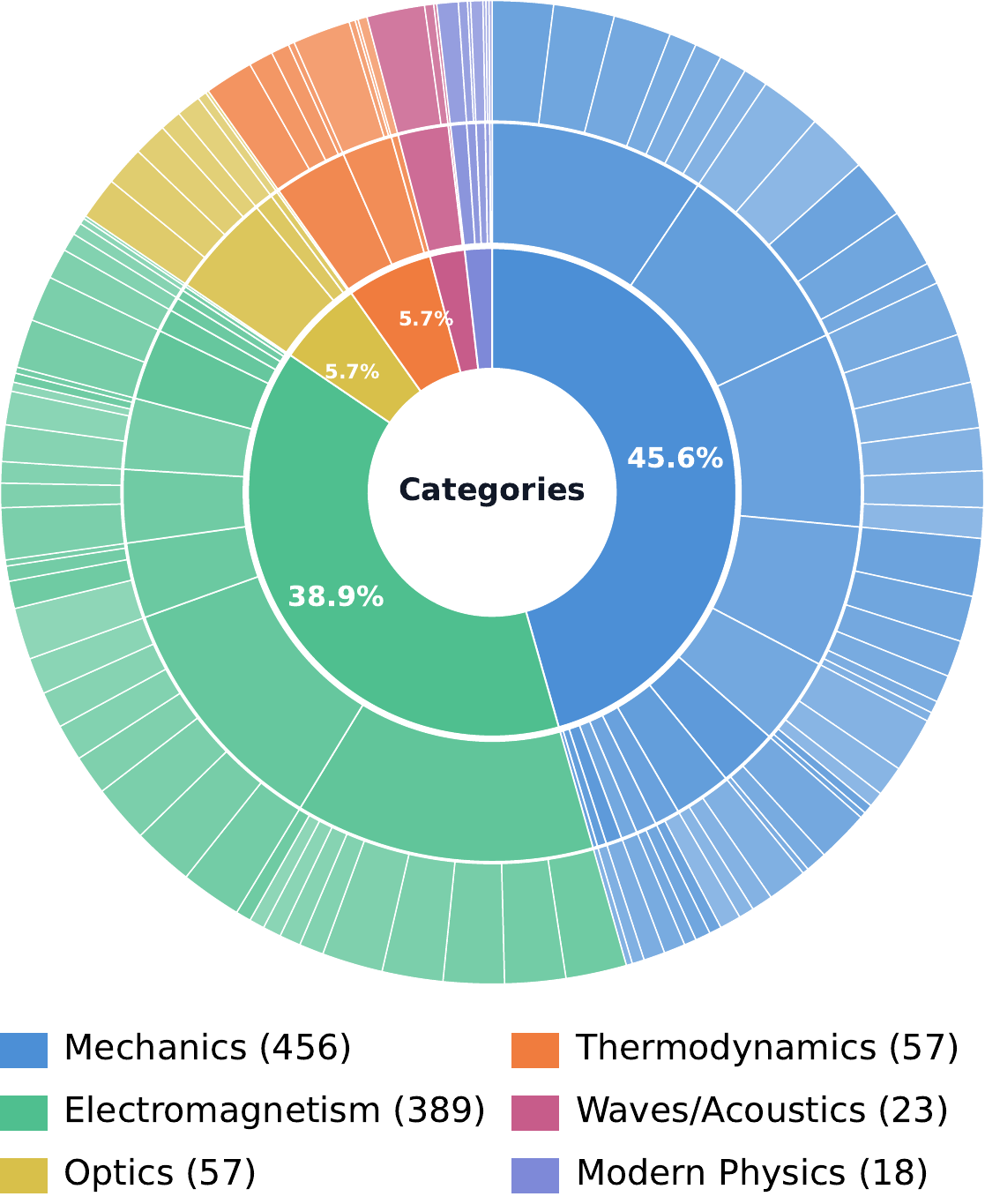}
    \end{minipage}
    \hfill
    \begin{minipage}[t]{0.28\textwidth}
        \centering
        \captionof{table}{\textbf{Dataset statistics.}}
        \label{tab:dataset_statistics}
        \vspace{0.35em}
        \scriptsize
        \begin{tabular}{lc}
            \toprule
            \textbf{Statistics} & \textbf{Number} \\ \midrule
            Seed Questions & 1,000 \\
            Total Questions & 4,000 \\ \midrule
            Disciplines & 6 \\ 
            Domains & 38 \\ 
            Fields & 104 \\ \midrule
            \textit{Answer Type} & \\
            \quad Value & 46.6\% \\
            \quad Option & 46.0\% \\ 
            \quad Equation & 6.2\% \\ 
            \quad Expression & 1.2\% \\ \bottomrule
        \end{tabular}
    \end{minipage}
    \hfill
    \begin{minipage}[t]{0.40\textwidth}
        \centering
        \captionof{table}{\textbf{Level-1-conditioned transfer accuracy.}}
        \label{tab:conditioned_transfer}
        \vspace{0.35em}
        \scriptsize
        \setlength{\tabcolsep}{2.4pt}
        \renewcommand{\arraystretch}{0.88}
        \begin{threeparttable}
        \begin{tabular}{lcccc}
            \toprule
            \textbf{Model} & $\boldsymbol{A_{1|1}}$ & $\boldsymbol{A_{2|1}}$ & $\boldsymbol{A_{3|1}}$ & $\boldsymbol{A_{4|1}}$ \\
            \midrule
            GPT-5.4 & 100.0 & 79.6 & 68.8 & 64.8 \\
            GPT-5 & 100.0 & 66.0 & 45.4 & 44.7 \\
            Gemini-3.1-Pro\tnote{a} & 100.0 & 88.7 & 81.0 & 77.5 \\
            Claude-4.7-Opus\tnote{a} & 100.0 & 81.8 & 68.2 & 57.4 \\
            Qwen3.5-27B & 100.0 & 55.6 & 46.4 & 41.0 \\
            Qwen-3.5-9B & 100.0 & 75.2 & 64.1 & 62.0 \\
            Gemma-4-26B-A4B-it & 100.0 & 54.1 & 44.9 & 37.3 \\
            Gemma-4-31B-it & 100.0 & 60.1 & 41.5 & 38.9 \\
            \bottomrule
        \end{tabular}
        \begin{tablenotes}[flushleft]
            \scriptsize
            \item \emph{Note.} $A_{\ell\mid 1}=100|\mathcal{C}_1\cap\mathcal{C}_\ell|/|\mathcal{C}_1|$. Row sizes are 539/334/142/148/360/242/292/311.
            \item[a] Evaluated on \texttt{testmini}.
        \end{tablenotes}
        \end{threeparttable}
    \end{minipage}
\end{figure*}

\paragraph{Taxonomy and metadata.}
\ourdataset is annotated with physics domains, visual information types, and reasoning skills.
The domain taxonomy covers major areas such as mechanics, electricity and magnetism, optics, thermodynamics, waves, and modern physics, with a long tail of more specialized topics, as summarized in Figure~\ref{fig:taxonomy_distribution}; Table~\ref{tab:dataset_statistics} reports the corresponding dataset scale and answer-type distribution.
Visual evidence is categorized by the type of information required for solution, including structure/topology, variable labels, directions and vectors, graphs and curves, geometric relations, and symbolic diagrams.
Reasoning metadata further records skills such as conservation-law reasoning, force analysis, circuit reduction, graph-to-equation conversion, geometric optics, unit reasoning, and multi-step numerical derivation.
These annotations support fine-grained analysis of whether failures arise from structural perception, variable grounding, or downstream physical reasoning.

\paragraph{Diagnostic metrics.}
For a model $f$ and an aligned test set
$\mathcal{D}=\{(x_i^{(1)},x_i^{(2)},x_i^{(3)},x_i^{(4)},y_i)\}_{i=1}^{N}$,
where $x_i^{(\ell)}$ is the Level-$\ell$ representation of the same seed problem, we define the Level-$\ell$ accuracy as
\begin{equation}
A_{\ell}(f)=\frac{100}{N}\sum_{i=1}^{N}
\mathbb{I}\!\left[g\!\left(f(x_i^{(\ell)})\right)=y_i\right],
\end{equation}
where $g(\cdot)$ denotes answer extraction and normalization.
The modality-transfer gaps are signed differences in percentage points:
\begin{align}
\Delta_{\mathrm{S}}(f) &= A_{1}(f)-A_{2}(f), \qquad
\Delta_{\mathrm{V}}(f) = A_{2}(f)-A_{3}(f), \nonumber\\
\Delta_{\mathrm{R}}(f) &= A_{3}(f)-A_{4}(f), \qquad
\Delta_{\mathrm{T}}(f) = A_{1}(f)-A_{4}(f).
\end{align}
Here $\Delta_{\mathrm{S}}$ measures the cost of transferring structural information into vision,
$\Delta_{\mathrm{V}}$ measures the additional cost of visually grounding variables,
$\Delta_{\mathrm{R}}$ measures the cost of full visual rendering, and
$\Delta_{\mathrm{T}}=\Delta_{\mathrm{S}}+\Delta_{\mathrm{V}}+\Delta_{\mathrm{R}}$ is the total transfer gap.
Positive values indicate degradation under a more visual representation, while negative values indicate that the visually richer variant is solved more accurately.
We also report four-way representation consistency,
\begin{equation}
\mathrm{Cons}_4 = \frac{100}{N}\sum_{i=1}^{N}\mathbb{I}
\left[
\hat{y}_{i,L1}=y_i \land \hat{y}_{i,L2}=y_i \land \hat{y}_{i,L3}=y_i \land \hat{y}_{i,L4}=y_i
\right],
\end{equation}
which measures the percentage of seed problems answered correctly across all four aligned representations.
Together, these metrics separate absolute problem-solving ability from robustness to visual modality transfer.

\section{Test-Time Modality Transfer}
\label{sec:evaluation}

This section focuses on the first research question of \ourdataset:
\emph{Can MLLMs maintain their performance when the same problem is expressed through progressively more visual and less textual representations?} We first outline our evaluation setup and present our main results.

\paragraph{Models.}
We evaluate 10 closed-weight and 5 open-weight MLLMs. The closed-weight set includes GPT-5.4 and GPT-5~\cite{openai_gpt5_systemcard}, Gemini-3.1-Pro and Gemini-3-Pro~\cite{team2023gemini}, Claude-4.7-Opus and Claude-4.6-Opus~\cite{anthropic_claude4}, Kimi K2.5~\cite{moonshot_kimi_k2}, Qwen-3.6-flash and Qwen3.5-122B-A10B~\cite{Qwen3,Qwen3-VL}, and SuperNova~\cite{supernova_model}. The open-weight set includes Qwen3.5-27B and Qwen-3.5-9B~\cite{Qwen3,Qwen3-VL}, P1-VL-30B-A3B~\cite{p1_vl}, and Gemma-4-26B-A4B-it and Gemma-4-31B-it~\cite{gemma}.

\paragraph{Test sets.}
For efficient development and API-cost control, each level is split into an 800-example \texttt{test} set and a 200-example \texttt{testmini} set with an 8:2 ratio. Unless otherwise stated, reported results use \texttt{test}; models marked in Table~\ref{tab:seephyspro_main} are evaluated on \texttt{testmini}. We evaluate the same seed problems across Level~1--4, enabling direct measurement of representation sensitivity under controlled modality transfer.

\paragraph{Judging.}
Following the evaluation practice of SeePhys~\cite{SeePhys} and the LMMS-Eval toolkit~\cite{lmms_eval}, we implement a composite answer judge. It first applies deterministic extraction and matching rules, including boxed-answer parsing, multiple-choice option matching, numerical tolerance, symbolic normalization, and unit-aware comparison. For outputs not resolved by these rules, we use DeepSeek-V3.2~\cite{deepseek_v32} as a more robust LLM judge. Closed-weight models are evaluated with a 32K context window and temperature $0$, except GPT-family models where the official API constraints require temperature $1$. Open-weight models are evaluated with a 16K context window. Additional benchmark and judging details are given in Appendix~\ref{app:evaluation_results}.

\paragraph{Main Results}
\label{sec:main_results}

\begin{table*}[t]
  \caption{\textbf{Main results on \ourdataset.}
  We report accuracy (\%) on four controlled modality-transfer levels, four-way representation consistency, and signed transfer gaps in percentage points.
  $\Delta_{\mathrm{S}}$, $\Delta_{\mathrm{V}}$, and $\Delta_{\mathrm{R}}$ correspond to Level 1$\rightarrow$2 structural transfer, Level 2$\rightarrow$3 variable grounding, and Level 3$\rightarrow$4 rendering, respectively; $\Delta_{\mathrm{T}}=A_1-A_4$ is the total gap.
  $\mathrm{Cons}_4$ denotes the percentage of test problems answered correctly at all four levels. Larger positive gaps indicate stronger representation sensitivity.}
  \label{tab:seephyspro_main}
  \begin{center}
    \begin{scriptsize}
      \begin{threeparttable}
      \setlength{\tabcolsep}{3pt}
      \renewcommand{\arraystretch}{0.95}
      \resizebox{\linewidth}{!}{%
      \begin{tabular}{l|ccccc|rrrr}
        \toprule
        \textbf{Model}
        & \multicolumn{5}{c|}{\textbf{Accuracy / consistency}}
        & \multicolumn{4}{c}{\textbf{Transfer gap}} \\
        & \textbf{L1} & \textbf{L2} & \textbf{L3} & \textbf{L4} & $\boldsymbol{\mathrm{Cons}_4}\uparrow$
        & $\boldsymbol{\Delta_{\mathrm{S}}}\downarrow$
        & $\boldsymbol{\Delta_{\mathrm{V}}}\downarrow$
        & $\boldsymbol{\Delta_{\mathrm{R}}}\downarrow$
        & $\boldsymbol{\Delta_{\mathrm{T}}}\downarrow$ \\
        \midrule
        \rowcolor{black!6}
        Human Performance & 54.0 & 58.5 & 59.5 & 56.0 & 49.0 & -4.5 & -1.0 & 3.5 & -2.0 \\
        \midrule
        \rowcolor{black!8}
        \multicolumn{10}{l}{\textbf{Closed-weight Frontier Models}} \\
        GPT-5.4 & 67.4 & 64.1 & 55.8 & 53.0 & 32.6 & 3.3 & 8.3 & 2.8 & 14.4 \\
        GPT-5 & 41.8 & 32.9 & 23.8 & 23.2 & 8.9 & 8.9 & 9.1 & 0.5 & 18.5 \\
        Gemini-3.1-Pro\tnote{a} & 71.0 & \textbf{72.0} & \textbf{66.5} & \textbf{66.5} & \textbf{47.0} & -1.0 & 5.5 & 0.0 & 4.5 \\
        Gemini-3-Pro & 58.9 & 51.2 & 46.2 & 45.5 & 43.0 & 7.7 & 5.0 & 0.7 & 13.4 \\
        Claude-4.7-Opus\tnote{a} & \textbf{74.0} & 67.0 & 56.5 & 46.5 & 33.5 & 7.0 & 10.5 & 10 & 27.5 \\
        Claude-4.6-Opus & 58.5 & 55.3 & 45.4 & 30.0 & 19.0 & 3.2 & 9.9 & 15.4 & 28.5 \\
        Kimi K2.5\tnote{a} & 52.0 & 48.5 & 46.0 & 42.0 & 26.0 & 3.5 & 2.5 & 4.0 & 10.0 \\
        Qwen-3.6-flash & 61.4 & 59.3 & 49.9 & 48.4 & 29.9 & 2.1 & 9.4 & 1.5 & 13.0 \\
        Qwen3.5-122B-A10B\tnote{a} & 47.5 & 48.0 & 39.5 & 42.0 & 25.5 & -0.5 & 8.5 & -2.5 & 5.5 \\
        SuperNova\tnote{a} & 25.5 & 27.5 & 20.0 & 22.0 & 11.0 & -2.0 & 7.5 & -2.0 & 3.5 \\
        \midrule
        \rowcolor{black!8}
        \multicolumn{10}{l}{\textbf{Open-weight Models}} \\
        Qwen3.5-27B & 45.0 & 34.8 & 28.0 & 25.6 & 9.9 & 10.3 & 6.8 & 2.4 & 19.4 \\
        Qwen-3.5-9B & 30.3 & 47.8 & 37.8 & 35.4 & 12.8 & -17.5 & 10.0 & 2.4 & -5.1 \\
        P1-VL-30B-A3B & 29.0 & 20.9 & 15.1 & 14.9 & 4.3 & 8.1 & 5.8 & 0.2 & 14.1 \\
        Gemma-4-26B-A4B-it & 36.5 & 29.5 & 26.1 & 19.4 & 9.0 & 7.0 & 3.4 & 6.7 & 17.1 \\
        Gemma-4-31B-it & 38.9 & 33.5 & 23.9 & 22.0 & 8.9 & 5.4 & 9.6 & 1.9 & 16.9 \\
        \midrule
        \rowcolor{black!8}
        \textbf{Average} & 49.2 & 46.1 & 38.7 & 35.8 & 21.4 & 3.0 & 7.4 & 2.9 & 13.4 \\
        \bottomrule
      \end{tabular}}
      \begin{tablenotes}[flushleft]
        \footnotesize
        \item \parbox{\linewidth}{\textsuperscript{a} Evaluated on the 200-sample \texttt{testmini} subset due to API-budget constraints.}
      \end{tablenotes}
      \end{threeparttable}
    \end{scriptsize}
  \end{center}
  \vskip -0.1in
\end{table*}

Across all evaluated models, average accuracy decreases from 49.2\% at Level 1 to 35.8\% at Level 4, yielding an average total modality-transfer gap of 13.4 points.
The degradation is not limited to weaker models: GPT-5.4 drops from 67.4\% to 53.0\%, and Claude-4.7-Opus drops from 74.0\% to 46.5\%.
Gemini-3.1-Pro is the strongest model on Level 4, but still does not match its own Level-1 performance.
As a human reference, 100 Chinese high-school students achieve 54.0\%, 58.5\%, 59.5\%, and 56.0\% on the \texttt{testmini} subset from Level 1 to Level 4.
Several frontier models exceed this reference in marginal accuracy, but none matches the human group's four-way consistency.

\textbf{Variable grounding is the dominant bottleneck.}
The staged gaps in Table~\ref{tab:seephyspro_main} show that moving only structure into images causes a smaller model-average gap ($\Delta_{\mathrm{S}}=3.0$), while moving variables and labels into images causes the largest drop ($\Delta_{\mathrm{V}}=7.4$).
The final rendering stage adds a smaller but non-negligible gap ($\Delta_{\mathrm{R}}=2.9$), reflecting the additional burden of OCR, formula recognition, and layout understanding.
Thus, the central failure mode is often not simply recognizing the diagram, but reading the right visual quantities and binding them to the correct physical entities.

\textbf{Marginal accuracy overestimates cross-representation stability.}
For example, Claude-4.7-Opus achieves the highest Level-1 accuracy but only 33.5\% four-way consistency, and GPT-5.4 has 32.6\% consistency despite 67.4\% Level-1 accuracy.

The marginal gap $\Delta_{\mathrm{T}}=A_1-A_4$ mixes two effects: whether the model can solve the underlying physics at all, and whether it can preserve that solution when information is moved into vision.
Table~\ref{tab:conditioned_transfer} removes the first factor by conditioning on problems that each model already solves at Level 1.
The remaining drops are still large: among Level-1-correct problems, GPT-5.4 retains 64.8\% accuracy at Level 4 and Claude-4.7-Opus retains 57.4\%.
These conditioned results show that the modality-transfer gap is not merely a consequence of difficult physics questions; models often lose an already-demonstrated solution when the same information is represented visually.

We further analyze performance by physics discipline in Appendix~\ref{app:discipline_results}, and present representative error clusters and case studies in Appendices~\ref{app:error_type_clustering} and~\ref{app:case_study_examples}.
Across these analyses, the dominant trend remains the same: visually grounded variable use is difficult beyond any single physics category.

\section{Training-Time Diagnostic: Can RL Help Close the Modality Gap?}
\label{sec:training_diagnostic}

Section~\ref{sec:evaluation} shows an inference-time failure: models lose accuracy when the same physics is represented more visually. This section asks a different but directly connected question: if we train on vision-necessary multimodal data, does RL actually close the modality-transfer gaps defined by \ourdataset? We therefore evaluate training by both accuracy gain and gap dynamics. A visually grounded improvement should not merely raise final-answer accuracy; it should preferentially improve the visually demanding levels and reduce gaps such as $\Delta_{\mathrm{V}}$ and $\Delta_{\mathrm{T}}$.

\subsection{Diagnostic Setup}
\label{sec:diagnostic_setup}

In addition to the benchmark itself, we construct and release two physics RL training corpora, \textbf{PhysRL-38K} and \textbf{PhysRL-8K}. This is motivated by a practical gap: multimodal physics datasets for RLVR remain scarce compared with visual math~\cite{MathVista,MathVerse,PAPO}, even though physics entails rich multimodal representation structure, from circuit analysis to Feynman diagrams.
PhysRL-38K is an approximately 38K-example physics VQA collection built from the same source pool and data engine as \ourdataset, covering public datasets, textbooks, olympiad archives, and exam-style problems. The training corpora are source-matched to \ourdataset but instance-disjoint from all benchmark test sets. Unlike the benchmark split, PhysRL-38K is designed for scalable training rather than controlled evaluation: it does not require the full manual redrawing, four-level alignment, and fine-grained modality-transfer annotation used by \ourdataset. We further obtain \textbf{PhysRL-8K} by filtering PhysRL-38K with GPT-5-mini~\cite{openai_gpt5_systemcard} to retain approximately 8K vision-necessary examples. Both PhysRL-38K and PhysRL-8K will be released, and Appendix~\ref{app:physics_training_pool_validation} validates the larger pool through RL runs that improve multiple held-out physics benchmarks.

We fine-tune Qwen2.5-VL-7B-Instruct~\cite{Qwen2.5-VL} and Qwen3-VL-4B-Instruct~\cite{Qwen3-VL} with outcome-supervised RL on physics and math vision-necessary corpora. For physics, the main training corpus is PhysRL-8K. For math, we construct \textbf{ViRL39K-VN}, a 22K-example vision-necessary subset selected from ViRL39K~\cite{PAPO,VLRethinker} with GPT-5-mini filtering. We use matched validation suites for the two domains. For physics, we evaluate on unmasked \ourdataset Level 1--4 and on held-out physics benchmarks including SeePhys~\cite{SeePhys}, PhysReason~\cite{PhysReason}, OlympiadBench~\cite{OlympiadBench}, and PhyX~\cite{PhyX}. For math, following the PAPO evaluation setting~\cite{PAPO}, we evaluate on the vision-dependent split of MathVerse~\cite{MathVerse}, referred to simply as \textbf{MathVerse}, and on \textbf{MMK12 Test}~\cite{MMK12}. Unless otherwise stated, all validation images are kept unmasked, so blind-training performance is measured on normal visual inputs.

All runs use a math-style final-answer prompt and an answer-verification reward. For the main runs, we use GSPO-style token-level policy optimization~\cite{GSPO} with four rollouts per prompt, rollout temperature $1.0$, top-$p=1.0$, maximum prompt and response lengths of 4096 tokens, and AdamW with learning rate $10^{-6}$ and weight decay $0.01$. We use one PPO epoch per update, bfloat16 FSDP training, vLLM rollout serving~\cite{vllm}, and evaluate every five training iterations. Qwen3-VL-4B runs use rollout batch size 256 and validation batch size 1024; Qwen2.5-VL-7B uses a larger rollout batch size of 512. Exact launch scripts and dataset variants are provided in the supplementary material.

\paragraph{Normal vs. blind RL.}
We compare standard RL with original images (\textit{normal RL}) against a matched blind-training control in which all training images are replaced with black images (\textit{blind RL}), while the train/test splits, reward function, and all other training settings remain unchanged.
We report normal and blind gains as follows,
\begin{align}
\mathrm{Gain}_{\mathrm{normal}} &= A_{\mathrm{normal}}-A_{0}, &
\mathrm{Gain}_{\mathrm{blind}} &= A_{\mathrm{blind}}-A_{0},
\end{align}
and define the visually grounded residual and blind-gain ratio as
\begin{align}
\mathrm{GroundedResidual} &= A_{\mathrm{normal}}-A_{\mathrm{blind}}, &
\rho_{\mathrm{blind}} &= \frac{\mathrm{Gain}_{\mathrm{blind}}}{\mathrm{Gain}_{\mathrm{normal}}}.
\end{align}
We also track gap closure,
\begin{equation}
\mathrm{Closure}(\Delta)=\frac{\Delta_{0}-\Delta_{\mathrm{after}}}{\Delta_{0}},
\end{equation}
where $\Delta$ is one of the modality-transfer gaps from Section~\ref{sec:benchmark}. Positive closure means RL reduces the benchmark-defined failure mode.

\subsection{SeePhys Pro Reveals Accuracy Gains without Gap Closure}
\label{sec:seephyspro_training_result}

\begin{figure*}[t]
    \centering
    \includegraphics[width=0.98\textwidth]{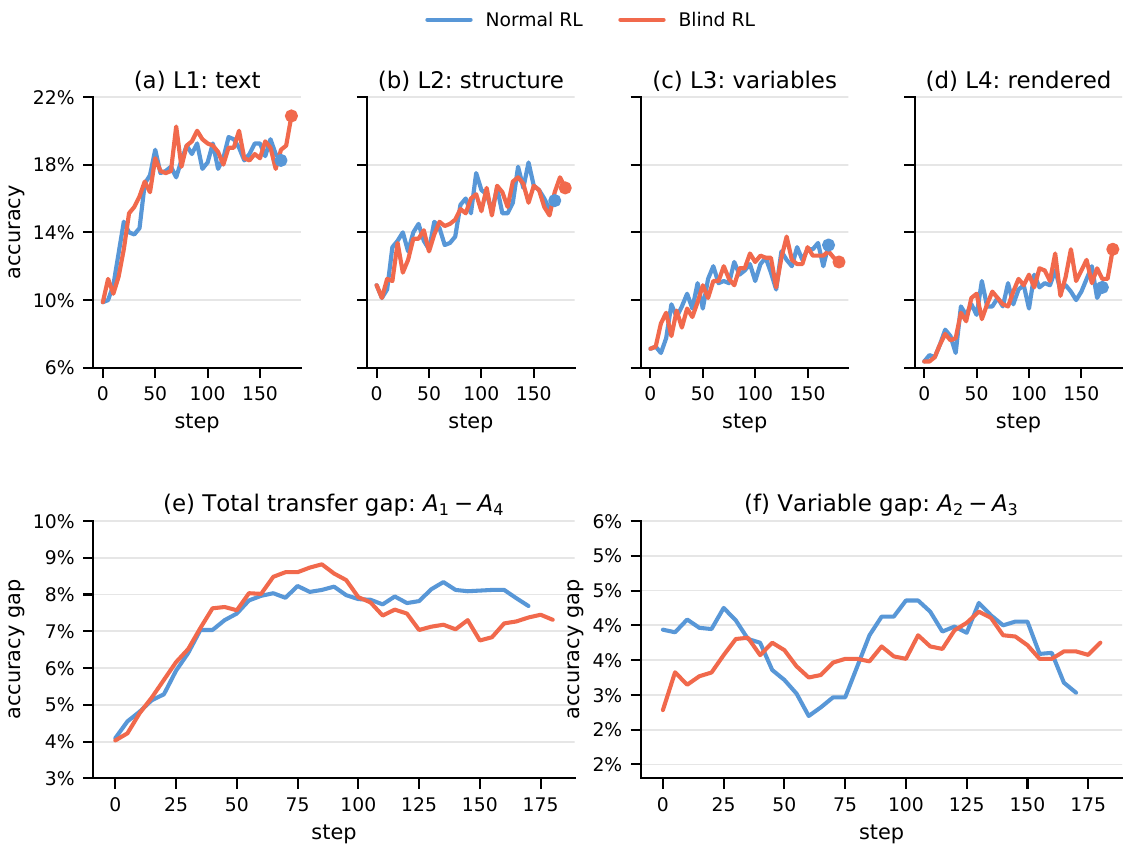}
    \vspace{-0.8em}
    \caption{\textbf{RL does not appear to close modality gaps.}
    Qwen3-VL-4B is trained on source-matched but test-disjoint vision-necessary physics data and evaluated on unmasked \ourdataset Level 1--4. The top row tracks validation accuracy on each modality-transfer level after normal and blind RL updates; the bottom row tracks the total transfer gap $\Delta_{\mathrm{T}}=A_1-A_4$ and variable-grounding gap $\Delta_{\mathrm{V}}=A_2-A_3$. Both normal and blind RL improve validation accuracy, but the visually induced gaps remain large.}
    \label{fig:train_diagnostic_combined}
    \vspace{-0.15em}
\end{figure*}

Figure~\ref{fig:train_diagnostic_combined} shows the central observation. Both normal and blind RL improve all four levels of \ourdataset. For Qwen3-VL-4B, Level-1 accuracy increases from $9.9\%$ to $18.3\%$ under normal RL and $20.9\%$ under blind RL; Level-4 accuracy increases from $6.4\%$ to $10.8\%$ and $13.0\%$, respectively.

The gap dynamics tell a different story. The total transfer gap $\Delta_{\mathrm{T}}$ widens from $3.5$ percentage points before training to $7.5$ points after normal RL and $7.9$ points after blind RL. The variable-grounding gap $\Delta_{\mathrm{V}}$ also does not show stable closure: the normal and blind curves in Figure~\ref{fig:train_diagnostic_combined}(f) remain close and even cross during training, rather than exhibiting a clear monotonic separation. 
The close trajectories of normal and blind RL indicate that valid training images are not sufficient by themselves to produce stable gap closure in this setting. Most of the gain appears to come from improvements in general physics problem solving, such as better physical-law selection, equation manipulation, numerical calculation, adaptation to common problem templates, and more reliable answer generation. These abilities can raise accuracy on all four levels, because the same underlying physics still has to be solved in every representation. By contrast, an improvement in visual grounding should make the model more reliable specifically when task-critical information is moved from text into the image, which would appear as larger gains on the more visual levels and a reduction in the modality gaps.

\subsection{Cross-Benchmark Negative Controls}
\label{sec:cross_benchmark_training}

\begin{figure*}[t]
    \centering
    \includegraphics[width=0.98\textwidth]{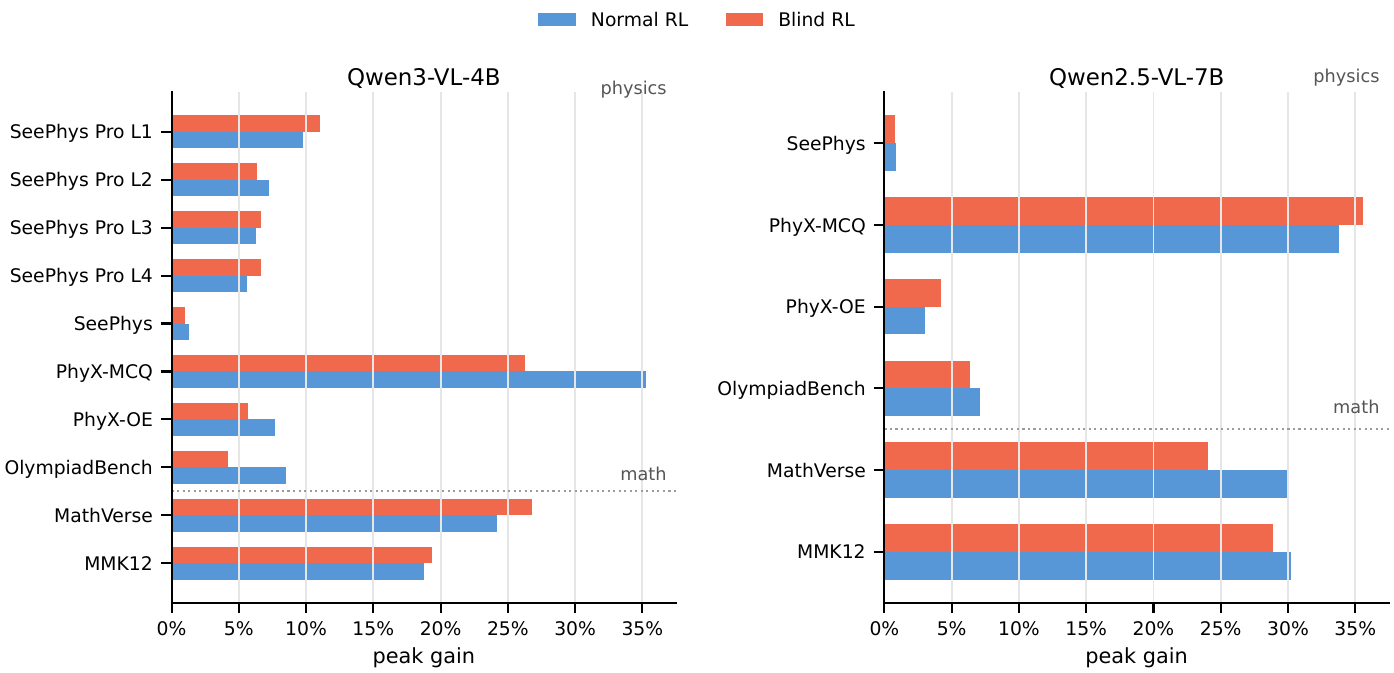}
    \vspace{-0.7em}
    \caption{\textbf{Blind gains are not unique to \ourdataset.}
    Peak validation gains for normal and blind RL on external physics and math benchmarks. Blind RL often recovers a substantial fraction of normal RL gains, and on several math settings it matches or exceeds normal RL.}
    \label{fig:cross_benchmark_training}
    \vspace{-0.8em}
\end{figure*}

Figure~\ref{fig:cross_benchmark_training} shows that blind-training gains are not unique to \ourdataset. The same negative control also produces gains on external math and physics benchmarks. The pattern is strongest on the math suite. On MathVerse and MMK12, blind RL often recovers a large fraction of the normal-RL gain, and in several settings reaches comparable or slightly higher peak gains. The physics suite is more heterogeneous. Normal RL is clearly stronger for Qwen3-VL-4B on benchmarks such as PhyX and OlympiadBench, suggesting that valid images can provide task-relevant signal in visually demanding physics problems. At the same time, the persistent blind gains across multiple physics evaluations show that outcome-RL on vision-necessary data can improve benchmark accuracy through non-visual adaptation; accuracy gains alone are therefore insufficient evidence of improved visual grounding. Additional details are reported in Appendix~\ref{app:training_diagnostic}.

\section{Mechanism Analysis: What Does Blind RL Learn?}
\label{sec:analysis_mitigation}

Blind RL improving on vision-necessary benchmarks is surprising, but it is not paradoxical. The key distinction is between instance-level vision necessity and distribution-level learnability. A single problem may be underdetermined without its image, yet a collection of such problems can still contain exploitable regularities in text, options, formula templates, answer ranges, and task style. We therefore analyze which non-visual signals remain available under blind training and which simple alternative explanations are insufficient.

\begin{figure*}[t]
    \centering
    \includegraphics[width=0.88\textwidth]{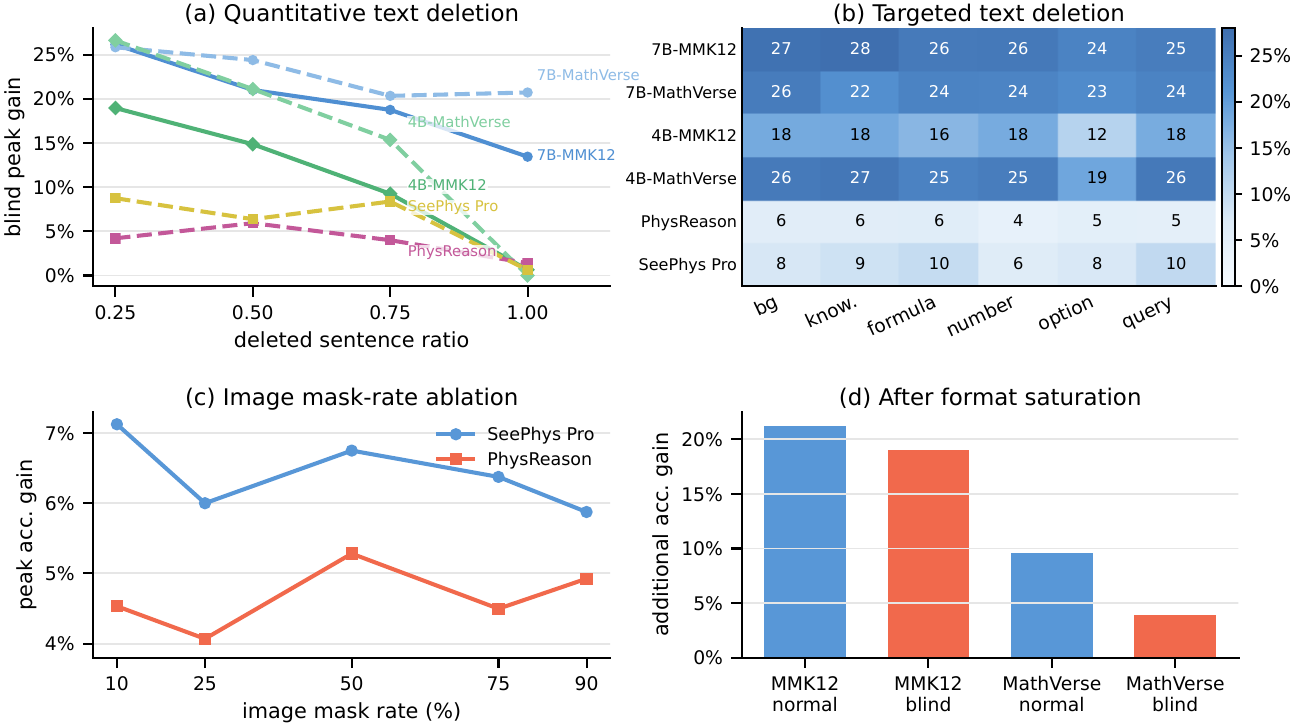}
    \vspace{-0.3em}
    \caption{\textbf{Mechanism controls for blind-training gains.}
    Text deletion, targeted deletion, mask-rate ablation, and post-format-saturation controls show that blind gains depend on residual textual/distributional cues rather than valid visual evidence. All gains are computed on unmasked validation sets.}
    \label{fig:mechanism_controls}
    \vspace{-0.2em}
\end{figure*}

\paragraph{Residual language is the main source of blind gains.}
Figure~\ref{fig:mechanism_controls}(a) shows that blind gains shrink as training text is deleted. For Qwen3-VL-4B, the blind peak gain on MathVerse~\cite{MathVerse} drops from about $26.6$ points at $25\%$ deletion to nearly zero at $100\%$ deletion; on MMK12~\cite{MMK12}, it drops from about $19.0$ points to below one point. Physics is noisier but shows the same boundary condition: complete text deletion reduces blind gains on PhysReason~\cite{PhysReason} and SeePhys Pro to near zero. Thus blind RL relies on residual language, problem style, answer priors, and generic reasoning practice rather than valid visual evidence.

\paragraph{The shortcuts are distributed rather than single-field artifacts.}
Figure~\ref{fig:mechanism_controls}(b) removes background, knowledge statements, formulas, numbers, options, and question clauses. In math, answer options are a visible shortcut source, especially for Qwen3-VL-4B on MMK12 and MathVerse. In physics, no single span explains the effect: deleting numbers, options, or formulas can reduce gains, but blind gains persist under most single-category deletions. The shortcut signal is therefore distributed across weak textual cues and dataset regularities.

\paragraph{Two auxiliary controls rule out simpler explanations.}
Figure~\ref{fig:mechanism_controls}(c) varies the training-image mask rate from $10\%$ to $90\%$. If blind gains came from a special all-black-image artifact, gains should change monotonically as masking increases. Instead, early peak gains remain positive and non-monotonic. Figure~\ref{fig:mechanism_controls}(d) also weakens the format-only explanation: after format reward reaches $90\%$, Qwen2.5-VL-7B still gains about $21.2/19.1$ points on MMK12 under normal/blind RL and $9.6/3.9$ points on MathVerse.

Together, these controls show that blind-training gains are non-visual gains induced by outcome-only RL on residual textual and structural regularities, explaining why Section~\ref{sec:training_diagnostic} observes accuracy improvement without proportional modality-gap closure. This is a credit-assignment failure in outcome-only multimodal RL~\cite{PAPO,VLRethinker}: final-answer rewards do not identify whether answers came from visual evidence, text priors, or shortcuts.

\paragraph{Limitations.}\label{sec:limitations}
\ourdataset is not intended to be the most difficult possible physics benchmark; its goal is controlled diagnosis of modality-transfer robustness rather than maximizing raw task difficulty. The paper also identifies inconsistency under modality transfer and non-visual gains under blind RL, but does not propose or validate a complete mitigation strategy. Promising directions include counterfactual image-text pairs, black-image unanswerability, and process-level rewards.

% Limitations are included as a paragraph in Section~\ref{sec:analysis_mitigation}
% to keep the main paper within the page budget.

\section{Conclusion}
\label{sec:conclusion}

We introduced \ourdataset, a progressive modality-transfer benchmark for multimodal physics reasoning.
By constructing four semantically aligned variants for each problem, \ourdataset diagnoses whether models preserve physical reasoning when information moves from text to visual structure, visual variables, and fully rendered diagrams.
Our results show that current MLLMs remain fragile under modality transfer, with visual variable grounding emerging as a key bottleneck.
We further developed PhysRL, a physics reasoning VQA training set for RL. Using them with \ourdataset as a diagnostic target, we find that blind training can improve validation accuracy even when training images contain no valid visual information. Together, these results motivate evaluating multimodal reasoning through both inference-time modality robustness and training-time grounding diagnostics, rather than relying only on final-answer accuracy.

\bibliography{Sections/custom}
\bibliographystyle{seephyspro_plainnat}
\newpage
\appendix
\section{Additional Evaluation Results}
\label{app:evaluation_results}

\subsection{Evaluation Protocol Details}
\label{app:evaluation_protocol}

All models are evaluated with the same answer-oriented prompt template. For Chinese problems, the instruction is the Chinese equivalent of asking the model to produce its reasoning as an internal monologue wrapped by \texttt{<thinking>} and \texttt{</thinking>} and then place the final answer in \texttt{\textbackslash boxed\{\}}. For English problems, we use the following instruction:
\begin{quote}
\small
\texttt{\{\{ content | trim \}\}}\\[0.3em]
\texttt{You first think through the reasoning process as an internal monologue, enclosed within <thinking> </thinking> tags. Then, provide your final answer enclosed within \textbackslash boxed\{\}.}\\
\end{quote}

For inference, closed-weight models use a 32K context window and temperature $0$ when supported. GPT-family models are evaluated with temperature $1$ due to official API constraints. Open-weight models use a 16K context window, with otherwise default greedy or deterministic decoding when available. Our judge follows the composite-evaluation style used in SeePhys~\cite{SeePhys} and LMMS-Eval~\cite{lmms_eval}. We first extract candidate final answers from \texttt{\textbackslash boxed\{\}} spans, final-answer markers, option letters, and trailing short answers. Deterministic rules then check multiple-choice options, normalized strings, symbolic expressions, numerical values under tolerance, and unit-aware equivalence. For unresolved or ambiguous cases, we call DeepSeek-V3.2~\cite{deepseek_v32} as an LLM judge with the problem, gold answer, and model prediction, and use it only to decide answer equivalence rather than to rescore reasoning quality.

\paragraph{Hardware.}
Local inference for open-weight models is run on NVIDIA GH200 GPUs. API-based closed-weight models are evaluated through their hosted providers. All RL experiments are conducted on 16 NVIDIA GH200 GPUs with bfloat16 FSDP training and vLLM rollout serving~\cite{vllm}.

\subsection{Benchmark Construction Details}
\label{app:construction_details}

\begin{figure*}[h]
    \centering
    \includegraphics[width=\textwidth]{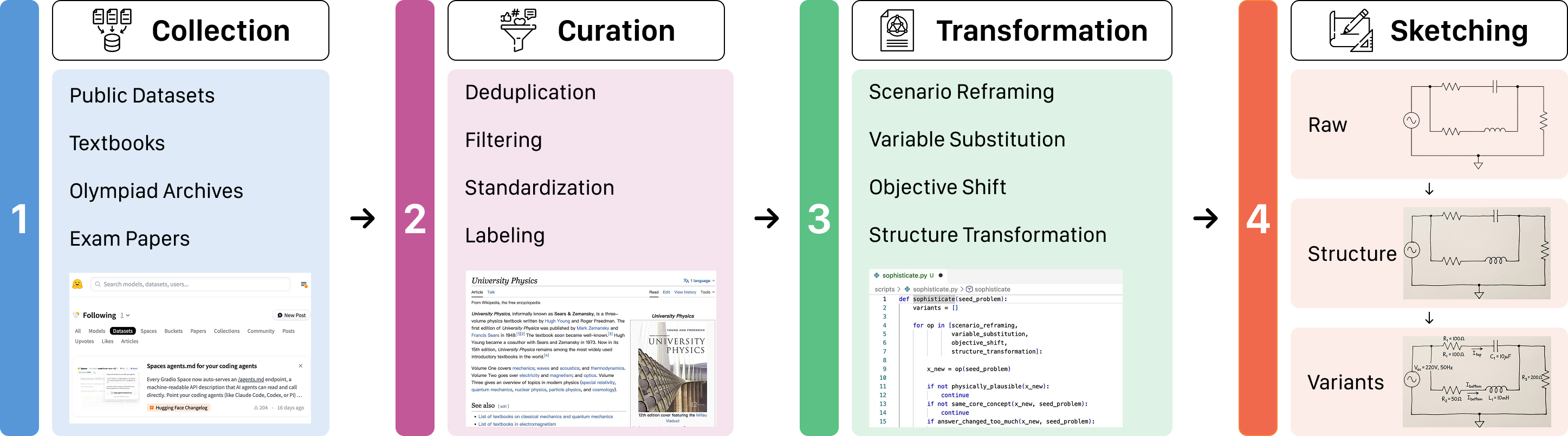}
    \caption{\textbf{Data pipeline for constructing \ourdataset.}
    We describe the construction process as four logical stages: source collection, curation, transformation, and controlled sketching into raw, structural, and variable-level variants.}
    \label{fig:app_data_engine}
\end{figure*}

We organize the benchmark construction pipeline into four logical stages: source collection, curation, transformation, and sketching, as summarized in Figure~\ref{fig:app_data_engine}. We first collect candidate problems from heterogeneous physics sources, including public datasets, textbooks and problem books (e.g., \emph{University Physics} and \emph{College Physics}), PhD qualifying and entrance examination collections, olympiad archives such as IPhO and CPhO, and school or university exam papers such as Cambridge IGCSE and AS/A-Level physics. The source pool contains over 5,000 PDF pages. Source PDFs are converted into structured text with Mathpix OCR~\cite{mathpix}, after which the curation stage performs deduplication, filtering, standardization, and labeling through text normalization, formula cleanup, and near-duplicate removal using exact-match checks, normalized edit-distance thresholds, and n-gram overlap signals, with manual review for formula-heavy borderline cases. In stages that require model assistance, including OCR cleanup, candidate normalization, and LLM-based verification of borderline cases, we use GPT-5-mini~\cite{openai_gpt5_systemcard}. The curated pool is then labeled by a team of ten annotators with engineering training (7 bachelor's-level and 3 PhD-level annotators). During this stage, each problem receives a three-level taxonomy covering discipline, field, and domain, together with auxiliary tags for required visual evidence and reasoning skills. The broad discipline layer includes major areas such as Classical Mechanics, Electromagnetism and Optics, and Statistical Mechanics and Thermodynamics, while lower field/domain layers capture more specific settings such as circuit analysis, rigid-body equilibrium, geometric optics, wave phenomena, and thermal processes.

After curation, accepted seeds are transformed into aligned four-level variants under the principle of \emph{same physics, different representation}. This transformation stage includes scenario reframing, variable substitution, objective shift, and structure transformation, while preserving the underlying physical system, target quantity, and gold answer. In the sketching stage, each problem is first hand-redrawn into a clean raw diagram that standardizes topology, geometry, arrows, and object boundaries. The visual content is then explicitly separated into a structure layer and a variable layer: the structure-only version retains entities and relations but removes symbolic quantities, whereas the variable layer overlays labels, values, and other quantity tokens onto the same diagram. This separation directly supports Level~2 and Level~3. Finally, the complete statement, formulas, and diagram are jointly rendered into a single image for Level~4, while keeping the semantic content aligned with the text-only Level~1 variant.

\subsection{Benchmark Embeddings}
\begin{figure*}[h]
    \centering
    \includegraphics[width=\textwidth]{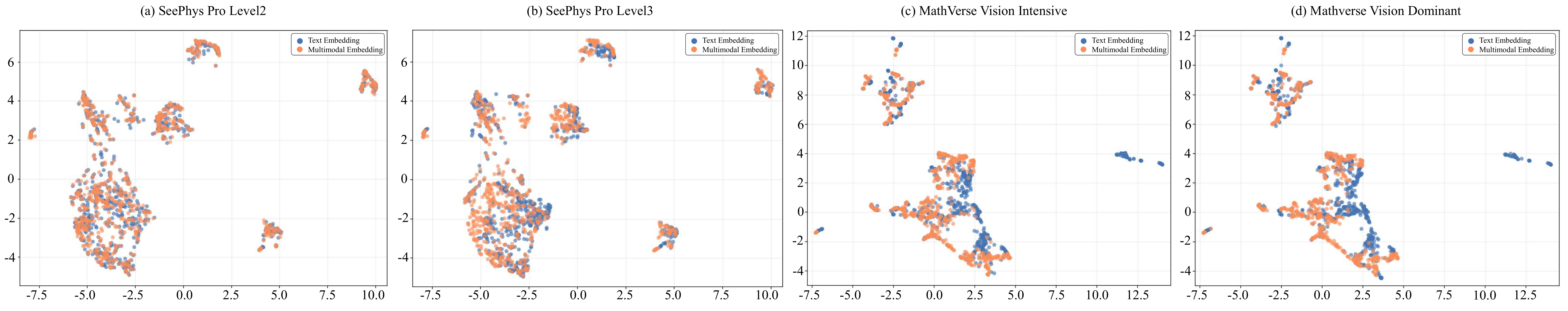}
    \caption{\textbf{\ourdataset and MathVerse Data Embeddings}
    Panels (a, b) are the embeddings of text and multimodal inputs at Level 2 and Level 3 of \ourdataset. Panels (c, d) are the embeddings of text and multimodal inputs for the Vision Intensive and Vision Dominant subsets of MathVerse.}
    \label{fig:data_embeddings}
\end{figure*}
Figure~\ref{fig:data_embeddings} provides a qualitative embedding visualization. In this view, the text-only and multimodal inputs in \ourdataset appear more separated at Level~3 than at Level~2, while the two MathVerse subsets appear broadly comparable. We use this figure as an illustrative distributional snapshot rather than as a quantitative metric.

\subsection{Discipline-Level Results}
\label{app:discipline_results}

Table~\ref{tab:discipline_l3} reports Level-3 accuracy across physics disciplines. Since Level 3 places both structural information and variables in the image, this setting directly evaluates visually grounded physics reasoning within each domain. The results should be interpreted primarily as a diagnostic breakdown rather than a fine-grained leaderboard. Mechanics and Electromagnetism dominate this split with 456 and 389 examples, respectively, while the remaining disciplines contain substantially fewer samples. Therefore, performance on Thermodynamics, Optics, Waves/Acoustics, and Modern Physics should be interpreted with caution.

\begin{table*}[t]
  \caption{\textbf{Level-3 accuracy by physics discipline.}}
  \label{tab:discipline_l3}
  \begin{center}
    \begin{small}
      \begin{threeparttable}
      \setlength{\tabcolsep}{4pt}
      \renewcommand{\arraystretch}{0.95}
      \begin{tabular}{l|cccccc}
        \toprule
        \textbf{Model}
        & \textbf{Mech.} & \textbf{EM} & \textbf{Thermo.} & \textbf{Optics} & \textbf{Waves} & \textbf{Mod.\ Phys.} \\
        \midrule
        \rowcolor{black!8}
        \textbf{\# Examples} & 456 & 389 & 57 & 57 & 23 & 18 \\
        \midrule
        GPT-5.4    & 60.5 & 50.8 & 53.1 & 47.9 & 93.3 & 38.5 \\
        GPT-5      & 23.2 & 23.6 & 24.5 & 25.0 & 40.0 & 15.4 \\
        Claude-Opus-4.7\tnote{a} & 56.4 & 59.2 & 50.0 & 44.4 & 62.5 & 40.0 \\
        Gemini-3.1-Pro\tnote{a}  & 69.1 & 67.1 & 62.5 & 44.4 & 62.5 & 60.0 \\
        Qwen3.6-Flash & 55.8 & 46.3 & 40.8 & 29.2 & 86.7 & 38.5 \\
        \midrule
        Qwen-3.5-9B   & 43.1 & 34.2 & 26.5 & 22.9 & 73.3 & 30.8 \\
        Qwen-3.5-27B  & 34.8 & 20.8 & 28.6 & 14.6 & 60.0 & 23.1 \\
        Gemma-4-26B-It & 29.3 & 23.0 & 22.5 & 22.9 & 40.0 & 23.1 \\
        Gemma-4-31B-It & 28.7 & 19.8 & 26.5 & 14.6 & 20.0 & 15.4 \\
        \bottomrule
      \end{tabular}
      \begin{tablenotes}[flushleft]
        \footnotesize
        \item Mech.: Mechanics; EM: Electromagnetism; Thermo.: Thermodynamics; Optics: Optics; Waves: Waves/Acoustics; Mod.\ Phys.: Modern Physics.
        \item[a] Evaluated on the 200-problem testmini subset.
      \end{tablenotes}
      \end{threeparttable}
    \end{small}
  \end{center}
  \vskip -0.1in
\end{table*}

\section{Additional Training Diagnostic Results}
\label{app:training_diagnostic}

\begin{figure*}[h]
    \centering
    \includegraphics[width=0.82\textwidth]{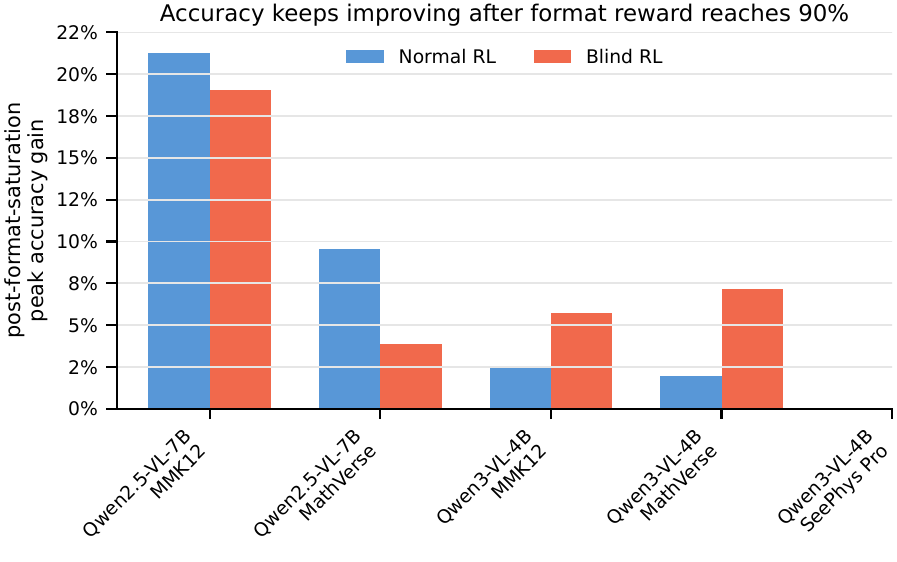}
    \caption{\textbf{Post-format-saturation gains.}
    After the format reward crosses $90\%$, answer accuracy can still increase substantially, especially in the Qwen2.5-VL-7B math runs highlighted in the main paper. This helps rule out a purely format-compliance explanation for blind gains.}
    \label{fig:app_post_format_gain}
\end{figure*}

\begin{figure*}[h]
    \centering
    \includegraphics[width=0.86\textwidth]{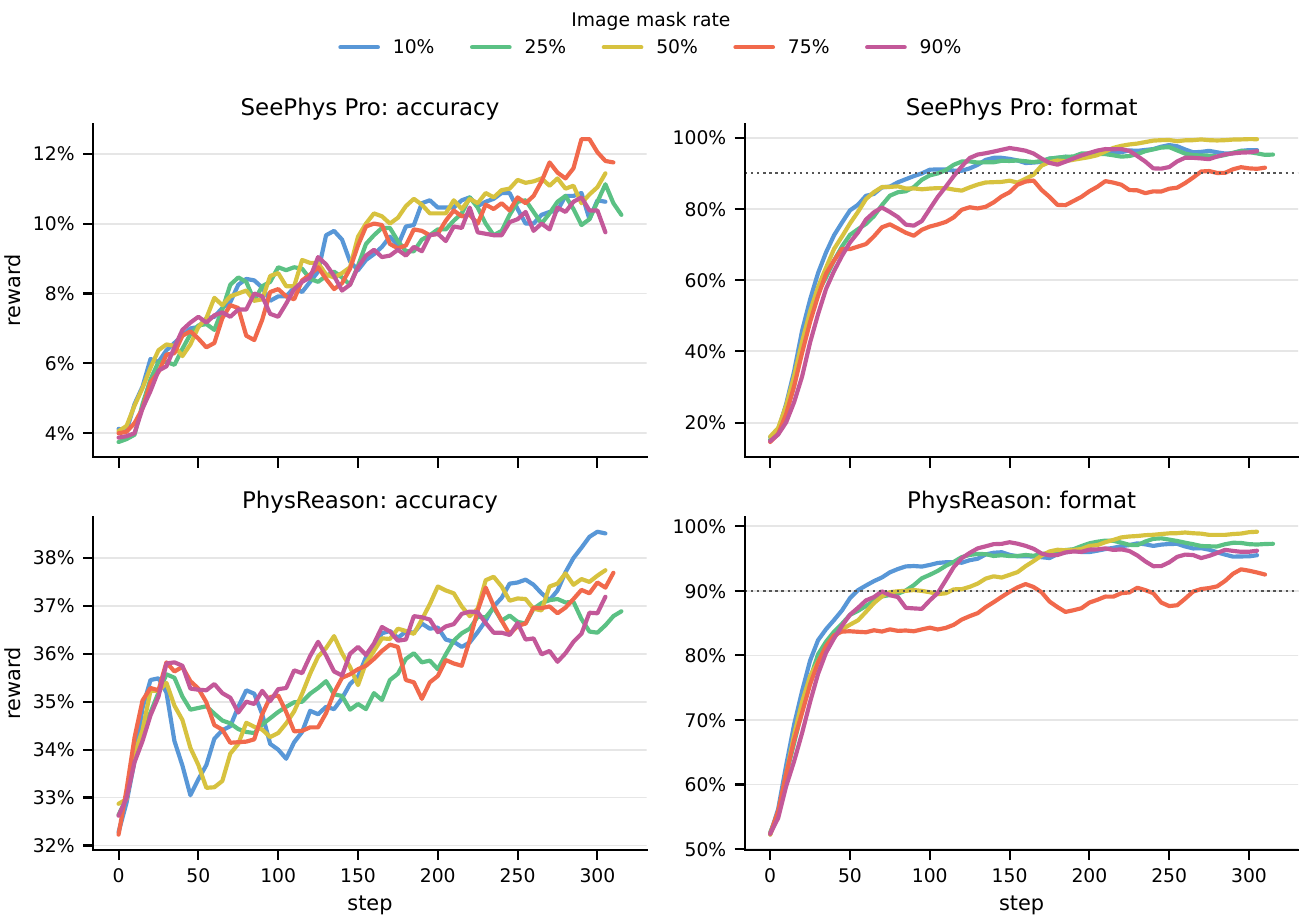}
    \caption{\textbf{Full image-mask-rate ablation curves.}
    We plot accuracy and format reward for Qwen3-VL-4B trained with different image mask rates and evaluated on unmasked SeePhys Pro and PhysReason. Main-paper Figure~\ref{fig:mechanism_controls} summarizes the peak gains within the first 200 updates.}
    \label{fig:app_image_mask_rate_curves}
\end{figure*}

\begin{figure*}[h]
    \centering
    \begin{minipage}[t]{0.49\textwidth}
        \centering
        \includegraphics[width=\linewidth]{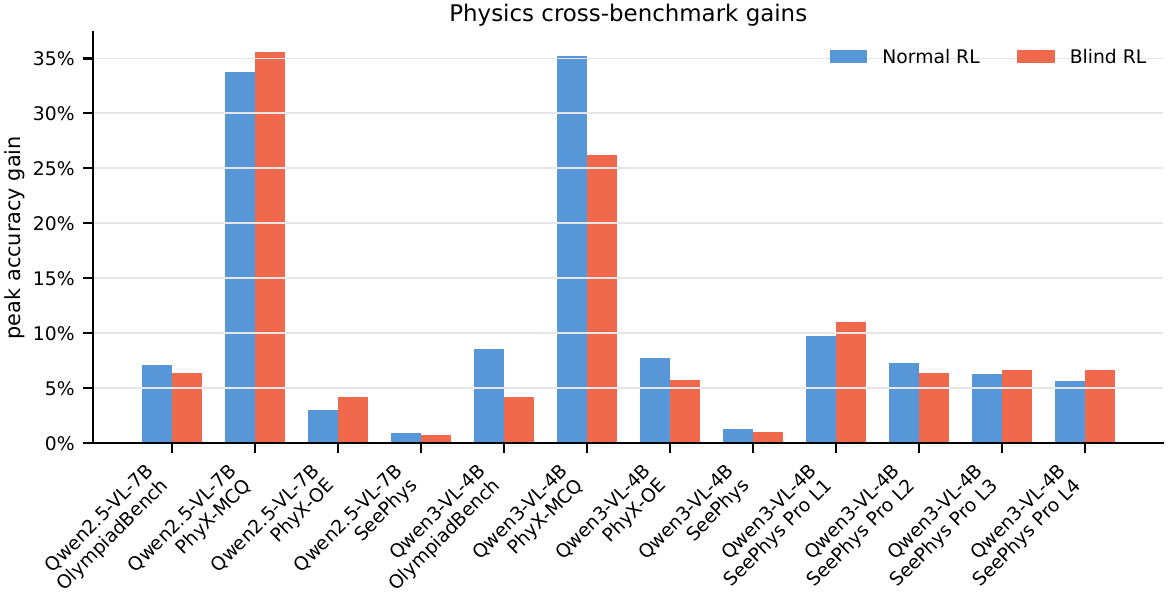}
    \end{minipage}
    \hfill
    \begin{minipage}[t]{0.49\textwidth}
        \centering
        \includegraphics[width=\linewidth]{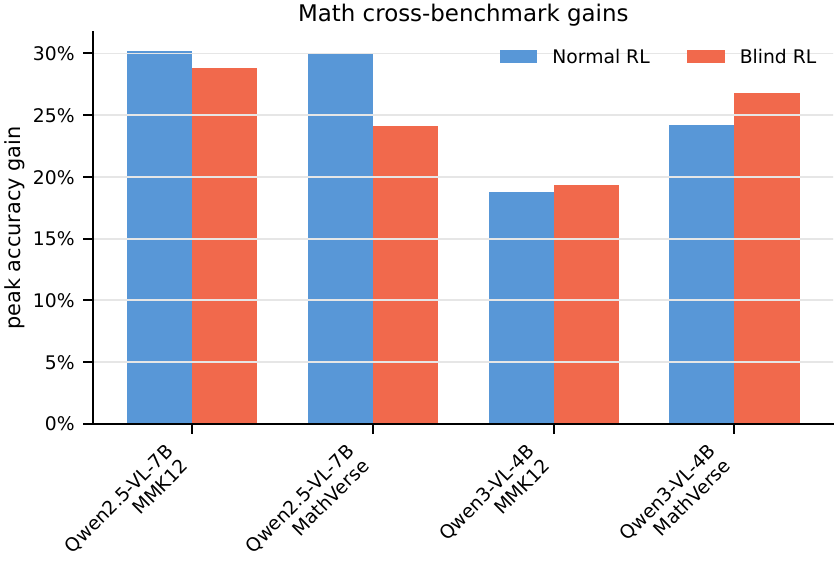}
    \end{minipage}
    \caption{\textbf{Separate cross-benchmark peak-gain plots.}
    These are expanded views of Figure~\ref{fig:cross_benchmark_training}.}
    \label{fig:app_cross_benchmark_separate}
\end{figure*}

This appendix reports additional visualizations generated from the SwanLab exports. Main-paper claims use smoothed curves for readability, but all summaries are computed from the raw exported validation records. Unless otherwise stated, RL runs use 16 NVIDIA GH200 GPUs. The appendix includes PhysRL-38K validation (Figure~\ref{fig:app_physrl38k_validation}), post-format-saturation gains (Figure~\ref{fig:app_post_format_gain}), full image-mask-rate curves (Figure~\ref{fig:app_image_mask_rate_curves}), and expanded cross-benchmark peak-gain plots (Figure~\ref{fig:app_cross_benchmark_separate}).

\subsection{Physics Training Pool Validation}
\label{app:physics_training_pool_validation}

\begin{figure*}[h]
    \centering
    \includegraphics[width=0.88\textwidth]{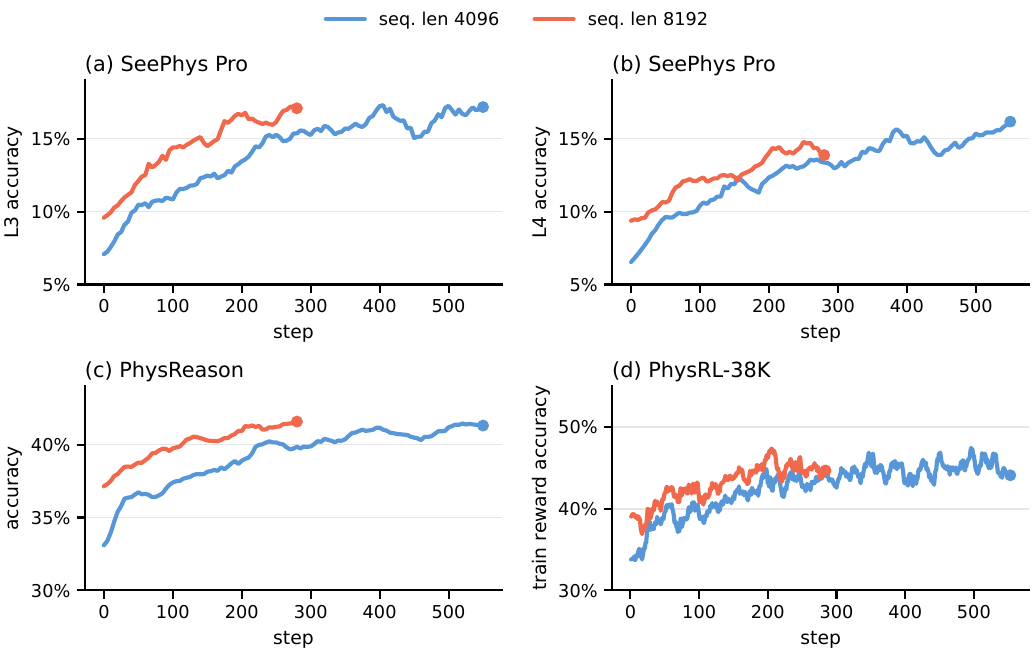}
    \caption{\textbf{PhysRL-38K training-pool validation with Qwen3-VL-4B.}
    We train with GSPO on PhysRL-38K using two sequence-length settings and evaluate on held-out, unmasked validation sets. Curves show that validation accuracy improves on SeePhys Pro Level~3/4 and PhysReason, while training reward accuracy also increases. Duplicate values from interrupted/resumed runs are averaged by step before smoothing for display.}
    \label{fig:app_physrl38k_validation}
\end{figure*}

PhysRL-38K is the larger source-matched physics training pool used to derive PhysRL-8K. The training examples are instance-disjoint from all benchmark test sets. As a quality check, we train Qwen3-VL-4B with GSPO on PhysRL-38K and evaluate on held-out physics validation sets, including SeePhys Pro Level~3/4 and PhysReason. Figure~\ref{fig:app_physrl38k_validation} shows two context-length settings. Both runs improve validation accuracy on SeePhys Pro and PhysReason while also increasing training reward accuracy, indicating that PhysRL-38K contains transferable physics reasoning signal rather than only \ourdataset-specific artifacts. Because some SwanLab exports contain resume-induced duplicate metric columns or repeated steps, we average duplicated values at the same step and apply light moving-average smoothing only for visualization. We include additional cross-benchmark diagnostic plots below; all summaries are computed from raw SwanLab validation exports.

% ============================================================
% Error clustering + case-study appendix layout.
%
% Required files:
%   figures/error_breakdown_3x4.pdf
%   figures/case_study_0428_tight_crop.pdf
%
% Notes:
% - The case-study PDF is already laid out as 6 pages.
% - This LaTeX block only inserts those existing PDF pages.
% - No internal text boxes, fonts, line spacing, or figure geometry inside the PDF are edited.
% - Error-Type Clustering explanatory text keeps line numbers.
% - The error-breakdown figure, figure caption, and case-study figures have no line numbers.
% ============================================================

\clearpage
\onecolumn

\begingroup
\setlength{\parindent}{0pt}
\setlength{\parskip}{0pt}
\setlength{\abovecaptionskip}{5pt}
\setlength{\belowcaptionskip}{0pt}

\ifcsname ErrorBreakdownW\endcsname\else\newlength{\ErrorBreakdownW}\fi
\setlength{\ErrorBreakdownW}{0.98\textwidth}

\ifcsname CasePageW\endcsname\else\newlength{\CasePageW}\fi
\setlength{\CasePageW}{0.90\paperwidth}

\ifcsname CasePairW\endcsname\else\newlength{\CasePairW}\fi
\setlength{\CasePairW}{0.74\paperwidth}

\ifcsname CaseCaptionW\endcsname\else\newlength{\CaseCaptionW}\fi
\setlength{\CaseCaptionW}{0.82\textwidth}

% If \captionof is unavailable, define a local fallback with the same text content.
\makeatletter
\ifcsname captionof\endcsname\else
  \newcommand{\captionof}[2]{%
    \refstepcounter{#1}%
    \par\smallskip\noindent\csname fnum@#1\endcsname: #2\par
  }%
\fi
\makeatother

\ifcsname CaseLocalCaption\endcsname
  \renewcommand{\CaseLocalCaption}[2]{%
    \vspace{0.45em}%
    \begin{minipage}{\CaseCaptionW}
      \footnotesize
      \captionof{figure}{#1}\label{#2}%
    \end{minipage}%
  }%
\else
  \newcommand{\CaseLocalCaption}[2]{%
    \vspace{0.45em}%
    \begin{minipage}{\CaseCaptionW}
      \footnotesize
      \captionof{figure}{#1}\label{#2}%
    \end{minipage}%
  }%
\fi

% First case-study page: no \clearpage here, and moved down from the section title.
\ifcsname CasePDFPageFigureFirst\endcsname
  \renewcommand{\CasePDFPageFigureFirst}[3]{%
    \thispagestyle{plain}%
    \vspace*{0.28cm}%
    \noindent\makebox[\textwidth][c]{%
      \begin{minipage}{\CasePageW}
        \centering
        \includegraphics[page=#1,width=\CasePageW,keepaspectratio]{figures/case_study_0428_tight_crop.pdf}\par
        \CaseLocalCaption{#2}{#3}%
      \end{minipage}%
    }\par
  }%
\else
  \newcommand{\CasePDFPageFigureFirst}[3]{%
    \thispagestyle{plain}%
    \vspace*{0.28cm}%
    \noindent\makebox[\textwidth][c]{%
      \begin{minipage}{\CasePageW}
        \centering
        \includegraphics[page=#1,width=\CasePageW,keepaspectratio]{figures/case_study_0428_tight_crop.pdf}\par
        \CaseLocalCaption{#2}{#3}%
      \end{minipage}%
    }\par
  }%
\fi

% A single case-study page after a page break, moved downward instead of upward.
\ifcsname CasePDFPageFigure\endcsname
  \renewcommand{\CasePDFPageFigure}[3]{%
    \clearpage
    \thispagestyle{plain}%
    \vspace*{0.28cm}%
    \noindent\makebox[\textwidth][c]{%
      \begin{minipage}{\CasePageW}
        \centering
        \includegraphics[page=#1,width=\CasePageW,keepaspectratio]{figures/case_study_0428_tight_crop.pdf}\par
        \CaseLocalCaption{#2}{#3}%
      \end{minipage}%
    }\par
  }%
\else
  \newcommand{\CasePDFPageFigure}[3]{%
    \clearpage
    \thispagestyle{plain}%
    \vspace*{0.28cm}%
    \noindent\makebox[\textwidth][c]{%
      \begin{minipage}{\CasePageW}
        \centering
        \includegraphics[page=#1,width=\CasePageW,keepaspectratio]{figures/case_study_0428_tight_crop.pdf}\par
        \CaseLocalCaption{#2}{#3}%
      \end{minipage}%
    }\par
  }%
\fi

% Two case-study PDF pages stacked on one output page.
% The two included images have exactly the same width: \CasePairW.
\ifcsname CaseTwoPDFPageFigures\endcsname
  \renewcommand{\CaseTwoPDFPageFigures}[6]{%
    \clearpage
    \thispagestyle{plain}%
    \vspace*{-0.45cm}%
    \noindent\makebox[\textwidth][c]{%
      \begin{minipage}{\CasePairW}
        \centering
        \includegraphics[page=#1,width=\CasePairW,keepaspectratio]{figures/case_study_0428_tight_crop.pdf}\par
        \CaseLocalCaption{#2}{#3}\par
        \vspace{0.00em}%
        \includegraphics[page=#4,width=\CasePairW,keepaspectratio]{figures/case_study_0428_tight_crop.pdf}\par
        \CaseLocalCaption{#5}{#6}%
      \end{minipage}%
    }\par
  }%
\else
  \newcommand{\CaseTwoPDFPageFigures}[6]{%
    \clearpage
    \thispagestyle{plain}%
    \vspace*{-0.45cm}%
    \noindent\makebox[\textwidth][c]{%
      \begin{minipage}{\CasePairW}
        \centering
        \includegraphics[page=#1,width=\CasePairW,keepaspectratio]{figures/case_study_0428_tight_crop.pdf}\par
        \CaseLocalCaption{#2}{#3}\par
        \vspace{0.00em}%
        \includegraphics[page=#4,width=\CasePairW,keepaspectratio]{figures/case_study_0428_tight_crop.pdf}\par
        \CaseLocalCaption{#5}{#6}%
      \end{minipage}%
    }\par
  }%
\fi

% ============================================================
% Error-type clustering before case studies.
% ============================================================

\section{Error-Type Clustering}
\label{app:error_type_clustering}

\vspace{0.4em}

% Keep line numbers only for the explanatory paragraph.
\ifcsname linenumbers\endcsname\linenumbers\fi
Before presenting individual case studies, we summarize the manually annotated
error clusters across modality-transfer levels and representative frontier models.
Figure~\ref{fig:error_breakdown_3x4} reports the distribution of error types for
GPT-5.4, Gemini-3.1-Pro, and Claude-Opus-4.7 from Level 1 to Level 4. The
cluster distribution shows that text-level failures are dominated by physics
modeling and reasoning errors, while visually richer settings introduce more
structural figure reading, numerical figure reading, and rendered-text reading
errors.
\par
\ifcsname nolinenumbers\endcsname\nolinenumbers\fi

\vspace{0.6em}

% The figure and its caption have no line numbers.
\thispagestyle{plain}%
\noindent\makebox[\textwidth][c]{%
  \begin{minipage}{\ErrorBreakdownW}
    \centering
    \includegraphics[width=\ErrorBreakdownW,keepaspectratio]{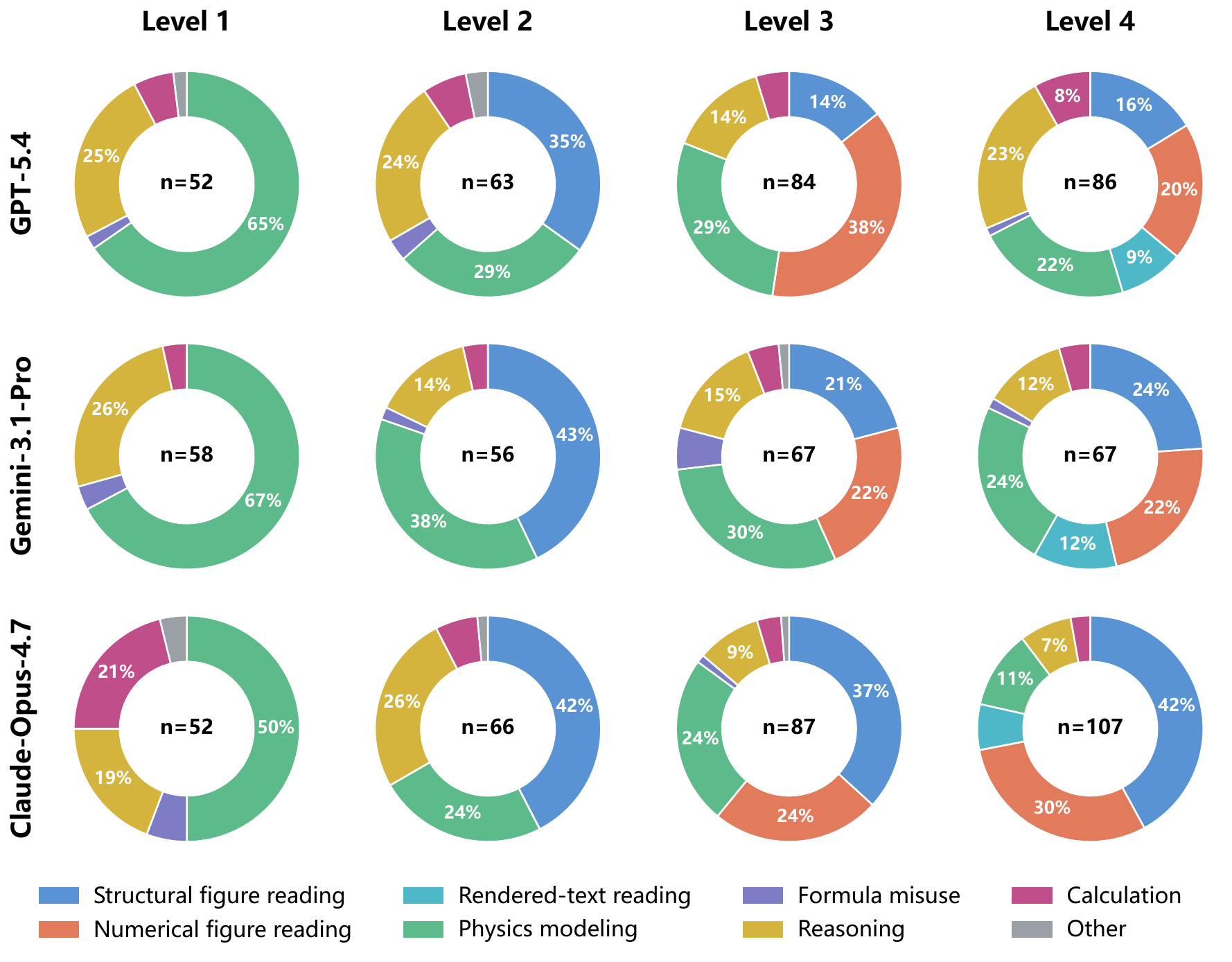}\par
    \vspace{0.20em}%
    \begin{minipage}{0.88\textwidth}
      \footnotesize
      \captionof{figure}{\textbf{Error-type clustering across modality-transfer levels and models.}
      Donut charts show the distribution of manually annotated error types for GPT-5.4,
      Gemini-3.1-Pro, and Claude-Opus-4.7 across Level 1--4. The center value $n$ denotes
      the number of analyzed errors for each model-level cell.}
      \label{fig:error_breakdown_3x4}%
    \end{minipage}%
  \end{minipage}%
}\par

\clearpage

% ============================================================
% Case-study examples. The case-study PDF is already laid out in 6 pages.
% ============================================================

% Give the Case Study section title a line number, but keep the figures unnumbered.
\ifcsname linenumbers\endcsname\linenumbers\fi
\section{Case Study Examples}
\label{app:case_study_examples}
\noindent The case studies below are selected independently for interpretability and are not restricted to the same fixed model set used in Figure~\ref{fig:error_breakdown_3x4}.
\par
\ifcsname nolinenumbers\endcsname\nolinenumbers\fi
\vspace{0.25cm}

\CasePDFPageFigureFirst
{1}
{\textbf{Oversimplification of physical modelling.}
Models oversimplified or misread the motion structure.}
{fig:case_group_force_structure}

\CasePDFPageFigure
{2}
{\textbf{Visual geometry grounding errors.}
Models misground geometric cues such as angles, radii, and arc marks.}
{fig:case_group_geometry_grounding}

\CaseTwoPDFPageFigures
{3}
{\textbf{Constraint and equilibrium failures.}
Models show constraint oversimplification, false equilibrium assumptions, and incorrect motion assumptions.}
{fig:case_constraint_equilibrium}
{4}
{\textbf{Transformer numerical misreading.}
Models misread numerical values in the visual input, which propagates into incorrect frequency and option judgments.}
{fig:case_transformer_numeric}

\CaseTwoPDFPageFigures
{5}
{\textbf{Reaction-force reasoning.}
The examples contrast correct force-balance reasoning with wrong reaction-direction and moment-equation assumptions.}
{fig:case_reaction_force}
{6}
{\textbf{Induced-charge calculation.}
The examples show how correct physical grounding can be disrupted by numerical misreading.}
{fig:case_induced_charge}

% Restore line numbers for the following text after this appendix block.
\ifcsname linenumbers\endcsname\linenumbers\fi
\endgroup

\clearpage

% \newpage
% \input{checklist}

\end{document}